\documentclass[a4paper,fleqn]{cas-sc}

\PassOptionsToPackage{table,xcdraw}{xcolor}

\usepackage[authoryear,longnamesfirst]{natbib}
\usepackage{algorithm}
\usepackage{graphicx}
\usepackage{subcaption}
\usepackage{bm,amsmath,amssymb,amsfonts}
\usepackage{times, array}
\usepackage{fancyhdr}
\usepackage{booktabs}
\usepackage[algo2e,ruled,lined]{algorithm2e} 

\begin{document}
\let\WriteBookmarks\relax
\def\floatpagepagefraction{1}
\def\textpagefraction{.001}

\shorttitle{OMG-RL: Offline Model-based Guided Reward Learning for Heparin Treatment}

\shortauthors{Lim and Lee}

\title [mode = title]{OMG-RL: Offline Model-based Guided Reward Learning for Heparin Treatment}                      
\tnotemark[1]

\tnotetext[1]{This work was supported by the National Research Foundation of Korea (NRF) under Grant 2021R1F1A1061093.}


\author[1]{Yooseok Lim}
\affiliation[1]{organization={Department of Industrial and Information Systems Engineering,
Soongsil University},
    city={Seoul},
    citysep={}, 
    postcode={06978}, 
    country={Republic of Korea}}
\ead{seook6853@soongsil.ac.kr}

\author[2]{Sujee Lee}[orcid=0000-0001-7511-2910]
\affiliation[2]{organization={Department of Systems Management Engineering,
Sungkyunkwan University},
    city={Suwon},
    citysep={}, 
    postcode={16419}, 
    country={Republic of Korea}}
\ead{sujeelee@skku.edu}

\cormark[1]
\cortext[cor1]{Corresponding author}

\begin{abstract}
Accurate medication dosing holds an important position in the overall patient therapeutic process. Therefore, much research has been conducted to develop optimal administration strategy based on Reinforcement learning (RL). However, Relying solely on a few explicitly defined reward functions makes it difficult to learn a treatment strategy that encompasses the diverse characteristics of various patients. Moreover, the multitude of drugs utilized in clinical practice makes it infeasible to construct a dedicated reward function for each medication. Here, we tried to develop a reward network that captures clinicians’ therapeutic intentions, departing from explicit rewards, and to derive an optimal heparin dosing policy.

In this study, we introduce Offline Model-based Guided Reward Learning (OMG-RL), which performs offline inverse RL (IRL). Through OMG-RL, we learn a parameterized reward function that captures the expert’s intentions from limited data, thereby enhancing the agent’s policy. We validate the proposed approach on the heparin dosing task. We show that OMG-RL policy is positively reinforced not only in terms of the learned reward network but also in activated partial thromboplastin time (aPTT), a key indicator for monitoring the effects of heparin. This means that the OMG-RL policy adequately reflects clinician’s intentions. This approach can be widely utilized not only for the heparin dosing problem but also for RL-based medication dosing tasks in general.
\end{abstract}

\begin{keywords}
inverse reinforcement learning \sep
offline reinforcement learning \sep
heparin dosing \sep
clinical decision support systems
\end{keywords}

\maketitle

\section{Introduction}
\label{sec:introduction}
Medication dosing is a crucial component of the patient treatment process. For instance, anticoagulants such as heparin and warfarin are widely used to prevent thrombosis \cite{1,2,3,4}, while propofol is administered to maintain stable conditions in anesthetized patients during surgical procedures \cite{5,6}. Precise chemotherapy is also vital for cancer patients \cite{7}. 
In medication dosing, key monitoring indicators play an essential role in ensuring appropriate dosage levels. For example, aPTT is used to adjust heparin, bispectral index and effect-side concentration guide anesthesia dosing, and cholesterol levels determine statin dosages \cite{3,4,5,8}.
These indicators are vital in guiding clinicians to administer medications accurately.

However, clinicians also consider emergency situations, comorbidities, genetic factors, and concurrent medications \cite{3,4}.
Heparin dosing guidelines, for instance, vary depending on the patient's specific condition, such as venous thromboembolism versus coronary artery disease \cite{3}. This reflects the complexity of determining appropriate medication dosages, which must consider various patient-specific factors and history. Therefore, a key objective in medication dosing is to derive logical dosages that encompasses multiple indicators. Implementing this comprehensive approach in dosing algorithms represents a significant advancement in the field.

RL provides a framework to derive personalized treatment policies by considering individual patient characteristics \cite{9}. Recently, RL has been applied to various medication dosing issues \cite{10,11,12,13}, with offline RL techniques becoming particularly notable for their effectiveness in settings where creating simulation environments is challenging \cite{10,11,12,13}. Notably, Xihe et al. \cite{12} used batch constrained Q-learning (BCQ) to optimize heparin dosing policies, while Smith et al. \cite{13} demonstrated the utility of conservative Q-Learning (CQL) in applying RL based on patient group characteristics. 

In such studies that utilize RL approaches, the precise definition of the Markov decision process (MDP) is essential for effective problem-solving.
Particularly, the reward function is a crucial element of the MDP in that it determines the learning direction of the target policy. Although previous approaches have often relied on specific clinical indicators such as aPTT to define rewards \cite{12,13,14,15}, it is clear that clinicians' decision-making processes consider a broader range of factors. This understanding indicates that reward functions in RL must be defined to reflect a more comprehensive set of variables. By incorporating diverse indicators that encompass both clinical and patient-specific factors, reward functions can better align with the complex decision-making processes of medical experts, thereby enhancing the efficacy and applicability of RL-based treatment strategies.

Considering these aspects, we adopt an inverse reinforcement learning (IRL) approach in this study \cite{16,17,18}. 
IRL, a category of imitation learning, learns a parameterized reward function that better captures the broad spectrum of expert behavior beyond single clinical indicators and utilizes it to evaluate and improve the agent's policy. 
This is crucial, as it aligns the learning process with real-world clinical decision-making that integrates various situational and patient-specific factors.
Furthermore, recognizing the challenges associated with traditional RL environments, we specifically focus on offline RL settings. 
These settings are pivotal in situations where real-time data collection or simulation is not feasible and only finite historical data is available. Recently, studies have been conducted in IRL to reflect the characteristics of off-policy learning and address offline problems \cite{50, 51, 52}, but in-depth research on application utilization is still needed.

To accommodate these constraints, we introduce Offline Model-based Guided Reward Learning (OMG-RL), a model-based IRL approach that effectively learns from limited data. 
OMG-RL is designed to increase the entropy of expert experience within the process of learning the reward function, thereby enhancing the robustness and applicability of the learned policies in real-world scenarios. Moreover, to handle the intricacies of state transitions in offline environments, OMG-RL incorporates a dynamic model capable of rollout, which facilitates better simulation and prediction capabilities.
We validate the proposed methodology using the heparin dosing problem. Our experimental data is sourced from MIMIC-III \cite{47}, a comprehensive public database containing de-identified health-related information from over 40,000 patients. The experiments demonstrate that our approach not only effectively learns the reward function from the data but also significantly improves the agent's policy implementation. This confirms the practical utility of our method in real-world settings.

The rest of this paper is organized as follows: Section II covers related work, and Section III outlines the theoretical background. Section IV details our methods and Section V and VI present the experimental setup and results. Finally, Section VII concludes with a summary and future direction.

\section{Related work}
\label{sec:related work}

\subsection{Heparin Treatment with RL}
Heparin administration studies favor offline RL approaches utilizing finite datasets due to the absence of simulators. Most studies on heparin treatment have focused on model-free RL approaches. These studies employ algorithms such as hidden Markov models (HMM), Q-learning, deep deterministic policy gradient (DDPG), BCQ, and CQL to learn optimal policies from finite datasets \cite{12,13,14,15,40,41}. 

Several studies have adopted model-based RL approaches \cite{28, 30} to enhance the performance of dosing policies. Baucum et al. \cite{42} proposed a strategy to address the heparin dosing problem using a transitional VAE to simulate the next state from the current patient state and physician action, effectively serving as a transition model. Further, Baucum et al. \cite{43} developed a method to classify patients into standard and non-standard groups for heparin dosing, learning tailored RL policies for the standard group and adapting these for the non-standard group, using the auto-regressive algorithm as the transition model.

previous researches require explicit reward functions derived from domain knowledge. IRL offers an effective alternative to address this challenge. While IRL utilization in medication dosing, including heparin treatment, remains limited. Similarly, Yu et al. \cite{44} applied IRL to sepsis treatment, building a reward function from three indicators related to sepsis and using random forests to find the optimal indicator combination for RL. This approach follows the traditional method of constructing reward functions from a linear combination of select indicators \cite{45}. Yu et al. \cite{46} used IRL for policies in intensive care units concerning mechanical ventilation and sedation, learning the weights of linear reward indicators through Bayesian fitted Q-iteration.

Despite a few IRL studies aiming to develop suitable reward functions beyond clinically defined indicators, these methods often still rely on extensive clinical knowledge. In contrast, our study estimates the reward function purely from clinicians' dosing experience data, avoiding predefined clinical indicators, and conducts both reward function estimation and RL policy learning in an offline setting.

\subsection{Offline Model-Based RL}
The offline model-based RL approach introduces a dynamic model to extend learning capabilities to the entire state-action space beyond the provided batch data, thereby enhancing the generalization of RL policies. This method utilizes supervised learning and generative modeling techniques as alternative strategies for policy learning. These techniques are especially useful in studies modeling complex, high-dimensional states such as those found in vision applications \cite{27}. Yu et al. \cite{28} proposed model-based offline policy optimization (MOPO), which effectively adapts the model-based approach for offline use. MOPO employs a dynamic model in a dyna-style configuration \cite{29}, quantifying the uncertainties to adjust the reward structure accordingly. Further extending this approach, Yu et al. \cite{30} introduced conservative offline model-based policy optimization (COMBO), which integrates CQL with MOPO to conservatively estimate the Q-function. COMBO penalizes out-of-distribution states generated during dynamic model simulations (rollouts), thereby leveraging the generalization advantages of model-based algorithms while avoiding the limitations imposed by uncertainty quantification. 

Here, we draw upon COMBO by leveraging dynamic models and conducting rollouts in offline settings to implement an expanded model-based RL strategy, thereby enhancing both the adaptability and efficacy of our approach.

\begin{figure*}[t!]
    \centering
    \includegraphics[width=1.0\linewidth]{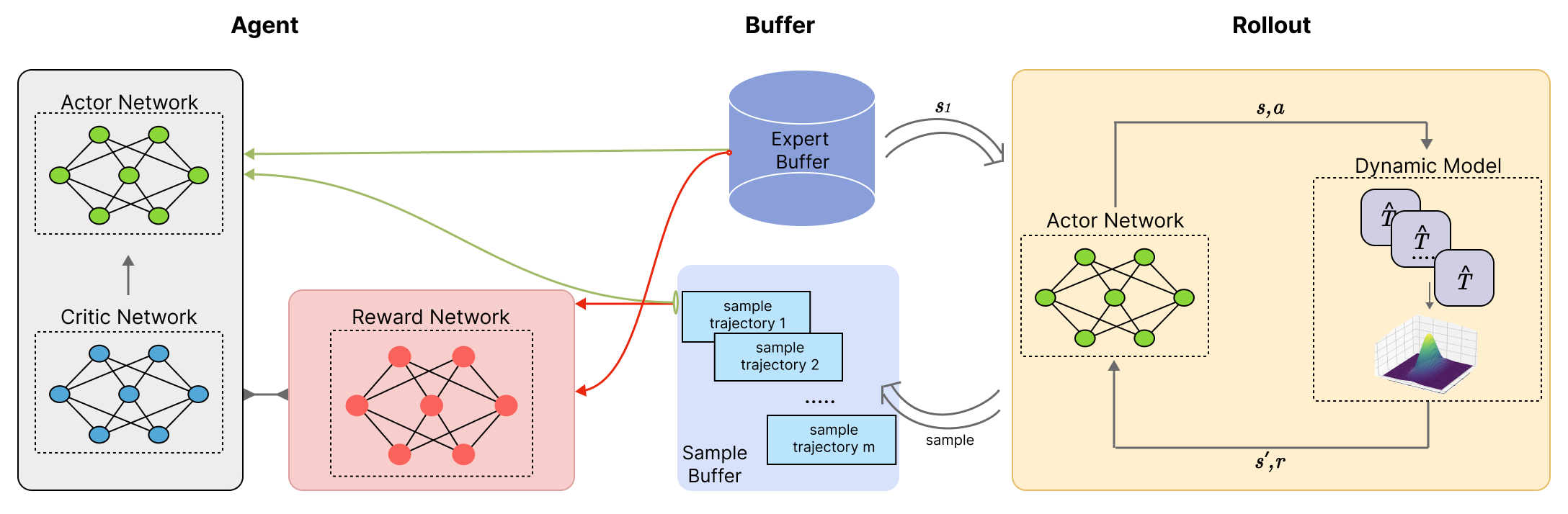}
    \caption{Diagram of the OMG-RL framework.}
    \label{fig: figure1}
\end{figure*}

\subsection{Online and Offline IRL}
IRL is a methodology that derives reward functions using expert trajectories \cite{16,17,18}. An expert trajectory, characterized by its demonstration of suboptimal yet effective outcomes, represents an experience of a policy that has satisfactorily achieved the problem's objective from a reinforcement learning perspective. The reward functions estimated through IRL are pivotal for learning optimal RL policies.

In the context of online IRL, the maximum entropy IRL (MaxEnt IRL) method \cite{31} validates the theory that maximizing entropy in alignment with the behavioral intentions of the expert trajectory for a specific policy facilitates the learning of expert behaviors. Wulfmeier et al. \cite{32} developed a method for performing maximum entropy estimation within the learning loop. However, this method encounters limitations due to its computational demands, as it requires alternating between reward updates and policy improvements through both external and internal loops. Finn et al. \cite{33} introduced a sample-based optimization method known as guided cost learning (GCL), which concurrently optimizes the reward function, built with neural networks, and the agent policy. More recently, Zeng et al. \cite{34} advanced IRL by adopting a maximum likelihood learning approach.

Regarding offline IRL, Klein et al. \cite{35} adapted classical apprenticeship learning (APP) \cite{17} to a batch and off-policy method by computing the expectation of features. Garg et al. \cite{36} introduced IQ-learn, which implicitly recreates rewards and policies using a learned soft Q-learning function. Additionally, Abdulhai et al. \cite{37} designed a novel approach that integrates multi-task RL pre-training with feature-based subsequent learning for performing IRL. Recently, various studies have also explored the complexities of offline IRL \cite{50, 51, 52}; however, limitations in the applicability of these approaches still exist. In response, this study proposes OMG-RL, a model-based offline IRL technique.

Specifically, we adopt the sample-based learning structure of GCL to estimate reward functions and perform IRL, utilizing a dynamic model to adapt it to an offline environment.

\section{Background}
\label{sec:background}
\subsection{Markov Decision Process (MDP)}
RL provides a framework for solving sequential decision-making problems, with the MDP serving as a fundamental problem definition in RL. An MDP is defined as a tuple \((\mathcal{S}, \mathcal{A}, r, \mathcal{P}, \gamma)\), where \(\mathcal{S}\) represents the set of states, \(\mathcal{A}\) the set of actions, \(\mathcal{P}\) the state transition probability \(P(s_{t+1}=s'|s_t=s, a_t=a)\), \(r: \mathcal{S} \times \mathcal{A} \rightarrow \mathcal{R}\) the reward function, and \(\gamma \in (0,1)\) the discount factor. The goal of an RL agent is to discover a policy \(\pi: \mathcal{S} \times \mathcal{A} \rightarrow (0,1)\) that maximizes the cumulative reward \(\sum_{t=0}^{\infty} \gamma^t r_t\), utilizing the trajectory \(\tau=(s_t, a_t, r_t)^T_{t=0}\).

\subsection{Maximum Entropy IRL}
Maximum entropy IRL \cite{31} aims to learn reward functions from expert demonstrations, utilizing an optimality variable $O$ to measure the effectiveness of a trajectory. 
The variable \(O_t\) serves as an indicator to evaluate optimality at state \(s_t\) and action \(a_t\) at time step \(t\), with the conditional probability defined as \(P(O_t | s_t, a_t) \propto \exp(r(s_t, a_t))\). Thus, the optimality of a trajectory adheres to the following equation:

\begin{equation}
\label{eq1}
 \begin{array}{cl}
    p(\tau|O_{1:T})=\frac{p(\tau,O_{1:T})}{p(o_{1:T})} \;\propto p(\tau)\prod_{t}\exp(r(s_t,a_t))
    \;=p(\tau)\exp(\sum_{t}r(s_t,a_t))
 \end{array}
\end{equation}

Given a set of trajectories \(\{\tau_i\}\) sampled from a policy \(\pi^*\) and a reward function \(r_\psi\), the reward function can be optimized by learning \(\psi\) in a direction that increases the likelihood of \({p(\tau|O_{1:T};\psi)}\). Maximum likelihood learning is performed according to the following equation:

\begin{equation}
\label{eq2}
    \max_{\psi}\frac{1}{N}\sum_{i=1}^{N}\log p(\tau_i|O_{1:T};\psi)=\max_{\psi}\sum_{i=1}^{N}r_{\psi}(\tau_{i})-\log Z
\end{equation}

The partition function \(Z=\int p(\tau) \exp(r_{\psi}(\tau))d\tau\) acts as a regularization term to prevent indiscriminate increase of rewards in expert trajectories. It uses the current policy's experience, encoded by \(r_\psi\), to modulate the degree of entropy maximization. The objective is to identify a \(\psi\) that assigns high rewards to expert trajectories, as reflected in the loss function:

\begin{equation}
\label{eq3}
 \begin{array}{cl}
    L=\frac{1}{N}\sum_{i=1}^{N}r_{\psi}(\tau_i)-\log\frac{1}{M}\sum_{j=1}^{M}p(\tau)\exp(r_{\psi}(\tau_j)) \approx E_{\tau \;\sim \pi^{*}(\tau)}[r_{\psi}(\tau)]-E_{\tau \sim p(\tau|O_{1:T;\psi})}[r_{\psi}(\tau)]
 \end{array}
\end{equation}

A set of trajectories \(\{\tau_j\}\) is sampled from the current policy. This loss function prioritizes higher rewards for the expert policy and lower rewards for others, effectively selecting the optimal policy based on the learned reward function. Moreover, when the goal is to identify a policy that encompasses the limited expert distribution while also considering the broader unexplored space, it effectively minimizes the likelihood of adopting a suboptimal policy. Ultimately, maximizing the entropy of a specific distribution serves to minimize the probability of selecting the least favorable policy.

\section{Methods}
\label{sec:methods}
 Fig. \ref{fig: figure1} illustrates the OMG-RL framework. Initially, we construct a dynamic model, a probabilistic model that describes patient state transitions and facilitates agent rollouts. Next, we undertake conservative policy evaluation and improvement to correct for state transition estimation errors from the dynamic model. Lastly, we guide the reward function to increase the entropy of the expert data and simultaneously perform reward function learning and policy updates within a singular learning loop.

\subsection{Dynamic Model}
The probabilistic dynamic model, denoted as \(\hat{T}(s',r \vert s,a)= \mathcal{N}\bigl(\mu_\theta(s, a), \Sigma_\theta(s, a)\bigr)\), takes state-action pairs \((s_t, a_t)\) as inputs and predicts the subsequent state and reward \((s_{t+1}, r_t)\) by sampling from a given distribution. From the perspective of RL, a probabilistic dynamic model effectively captures the uncertainty and variability of the environment, thereby enabling more robust and flexible policy learning compared to a deterministic model. Our approach implements an ensemble method. In model-based RL, using ensemble methods to construct \(\hat{T}\) significantly enhances performance \cite{28}. The networks are trained using maximum likelihood estimation as follows:

\begin{equation}
\label{eq4}
L=E_{(s,a,s',r) \sim D}[\log\hat{T}(s',r|s,a)]
\end{equation}

Employing a dyna-style approach \cite{29}, we integrate the dynamic model into RL. This technique uses augmented data for policy evaluation between iterative learning cycles. Data from original batch \(\mathcal{D}_{batch}\) and data obtained through rollouts using the current policy \(\mathcal{D}_{sample}\) are employed. In each iteration, an initial state \(s\) is randomly selected from \(\mathcal{D}_{batch}\), and rollouts of batch size \(b\) and length \(h\) are performed using \(\hat{T}\). The rollout data is then added to \(\mathcal{D}_{sample}\), and both datasets are used for policy evaluation and improvement. The procedure is detailed in Algorithm \ref{alg:alg1}.

\begin{algorithm}[H]
\caption{Dyna-style Algorithm}\label{alg:alg1}
\SetAlgoLined
\SetKwInput{KwRequire}{Require}
\SetNlSty{}{}{:}
\KwRequire{rollout horizon $h$, rollout batch size $b$, $D_{\text{batch}}$}

Train on batch data $D_{\text{batch}}$ an ensemble of $N$ probabilistic dynamic models $\{\hat{T}^i(s',r \vert s,a)\}_{i=1}^N$.

Initialize policy $\pi$ and an empty replay buffer $D_{\text{sample}} \leftarrow \emptyset$.

\For{epoch $= 1,2,\dots$}{
    \For{$i = 1$ to $b$ (in parallel)}{
        Sample initial state $s_1$ from $D_{\text{batch}}$.
        
        \For{$j = 1$ to $h$}{
            Sample an action $a_j \sim \pi(s_j)$.
            
            Randomly select a dynamics  $\hat{T}$ from $\{\hat{T}^i\}_{i=1}^N$.
            
            Sample $(s_{j+1}, r_j) \sim \hat{T}(s_j, a_j)$.
            
            Add $(s_j, a_j, r_j, s_{j+1})$ to $D_{\text{sample}}$.
        }
    }
    Draw samples from $D_{\text{batch}} \cup D_{\text{sample}}$.
    
    Update the policy.
}
\end{algorithm}

\subsection{Conservative Policy Evaluation and Improvement}
The dynamic model, trained on finite data, is prone to errors in estimating state transitions. Accurately estimating the uncertainty of these errors is crucial for achieving policy convergence. Following the COMBO approach \cite{30}, we use a conservative update method that accounts for the model's uncertainty during policy learning, optimizing the lower bound of policy performance. To ensure conservative policy updates, we penalize the Q-values for state-action pairs that likely deviate from the expected distribution and adjust Q-values for reliable pairs. The recursive update equation is as follows:

\begin{equation}
\label{eq5}
    \hat{Q}^{k+1} \leftarrow \arg\min_{Q} \;\alpha\left(E_{s,a \sim \rho(s,a)}[Q(s,a)] 
    - E_{s,a \sim \mathcal{D}_{batch}}[Q(s,a)]\right)
    +\frac{1}{2}E_{s,a,s' \sim d_{\lambda}}[(Q(s,a)-\hat{B}^{\pi}\hat{Q}^{\pi}(s,a))^2]
\end{equation}

Here, \(\alpha\) acts as the trade-off factor between the regularization term and the policy loss function. \(\hat{B}^{\pi}\) is the Bellman operator, ensuring that updates adhere to the principles of dynamic programming. The distribution \(\rho(s,a)=d^\pi_{\hat{T}}(s)\pi(a|s)\) is a sampling distribution from \(\hat{T}\). \(d^\pi_{\hat{T}}(s)\) is the discounted marginal state distribution when executing \(\pi\) under the \(\hat{T}\). For state-action pairs from \(\rho(s,a)\) that are out-of-distribution, we minimize their Q-function values to prevent overfitting to potentially erroneous data. Conversely, for state-action pairs from the trusted data in \(\mathcal{D}_{batch}\), we maximize their Q-function values, reinforcing the reliability of learned policies. The distribution \(d_{\lambda} = d_{\lambda}^{\mu}(s,a) := \lambda d_{batch} + (1- \lambda) d_{\hat{T}}^{\mu}(s,a)\), which includes both \(\mathcal{D}_{batch}\) and \(\mathcal{D}_{sample}\), balances the data used for updating Q-values according to a specified ratio \(\lambda \in [0,1]\), integrating both empirical and simulated data for a comprehensive policy evaluation. Following the estimation of conservative Q-values, policy improvement is executed as specified in (\ref{eq6}):

\begin{equation}
\label{eq6}
    \pi^* \leftarrow \arg\max_{\pi}E_{s \sim \rho, a \sim \pi(\cdot|s)}[\hat{Q}^{\pi}(s,a)]
\end{equation}

\subsection{Guided Reward}
To utilize a universal reward function for policy optimization, we develop a medication dosing policy based on GCL \cite{33}. We employ a sample-based methodology to learn in a direction that maximizes the entropy of the expert trajectory. The sample trajectory generated by the dynamic model is used to enhance the agent's policy and estimate the partition function of the reward function. Through policy improvement, the sampling distribution is refined to better estimate the corresponding distribution.

In MaxEnt IRL, designing the sampling distribution \(\pi(\tau)\) for the partition function \(Z\) is critical for convergence. Without specific information about the reward function, it is challenging to define \(Z\) as a particular distribution. Instead of fixing a specific distribution, \(\pi(\tau)\) is iteratively improved to generate more samples from significant regions of the trajectory space according to the current policy. The sample trajectory generated following \(\pi\) is then used to optimize the reward function. The optimization of the reward function and the policy function are conducted simultaneously to reflect the improvements in \(\pi(\tau)\). The loss function that incorporates this sampling optimization method for estimating the partition function is as follows:

\begin{equation}
\label{eq7}
 \begin{array}{cl}
    L&\approx\frac{1}{N}\sum_{i=1}^N r_{\psi}(\tau_i) - \frac{1}{M}\sum_{j=1}^M r_{\psi}(\tau_j) \\
    \\
    \nabla_{\psi}L &\approx\frac{1}{N}\sum_{i=1}^N \nabla_{\psi} r_{\psi}(\tau_i) - \frac{1}{M}\sum_{j=1}^M \nabla_{\psi} r_{\psi}(\tau_j)
 \end{array}
\end{equation}

Here, \(i\) represents the expert trajectory, \(j\) represents the sample trajectory, and \({\tau}_j\) is the data sampled from the \(\pi({\tau}_j)\). Due to the sample-based estimation of the partition function, estimates might be biased or distribute incorrectly. To correct this, importance sampling (IS) is employed to ensure consistent likelihood and reward function estimation. The importance weight is given by \(w_j = \frac{p(\tau) \exp(r_{\psi}(\tau_j))}{\pi(\tau_j)} = \frac{\exp(\sum_t r_{\psi}(s_t,a_t))}{\prod_t \pi(a_t|s_t)}\), and the objective function reflecting this is provided in the following equation:

\begin{equation}
\label{eq8}
    \nabla_{\psi}L \approx \frac{1}{N}\sum_{i=1}^N \nabla_{\psi} r_{\psi}(\tau_i) - \frac{1}{\sum_j w_j}\sum_{j=1}^M w_j \nabla_{\psi} r_{\psi}(\tau_j)
\end{equation}

In conclusion, the final methodology combines the dynamic model, conservative policy improvement, and guided reward approach as outlined in Algorithm \ref{alg:alg2}. The expert trajectory \(\mathcal{D}_{expert}\) guides initial training. After training the dynamic model using offline data, the policy initialization step is performed, and \(\mathcal{D}_{sample}\) is generated through the rollout of the dynamic model with \(\pi\). The reward function is optimized using both \(\mathcal{D}_{expert}\) and \(\mathcal{D}_{sample}\). If the batch size included in the initial updates is small, it may hinder learning convergence; hence, the expert trajectory is integrated into the sample trajectory. Finally, the Q-function is learned using the collected samples, and a conservative policy function update is executed.

\begin{algorithm}[]
\caption{Offline Model-Based Guided Reward Learning}\label{alg:alg2}
\SetAlgoLined
\SetKwInput{KwRequire}{Require}
\SetNlSty{}{}{:}
\KwRequire{rollout horizon \( h \), rollout batch size \( b \), \( D_{\text{expert}} \)}

Train on batch data \( D_{\text{expert}} \) an ensemble of \( N \) probabilistic dynamic models \(\{\hat{T}^i(s', r \mid s, a)\}_{i=1}^N\).

Initialize policy \( \pi_{\phi} \), critic \( Q_{\psi} \), and an empty replay buffer \( D_{\text{sample}} \leftarrow \emptyset \).

\For{epoch \( = 1,2,\ldots \)}{
  \For{\( i = 1 \) to \( b \) (in parallel)}{
    Sample initial state \( s_1 \) from \( D_{\text{expert}} \).
    
    \For{\( j = 1 \) to \( h \)}{
      Sample an action \( a_j \sim \pi_{\phi}(s_j) \).
      
      Randomly select a dynamics \( \hat{T} \) from \(\{\hat{T}^i\}\).
      
      Sample \( (s_{j+1}, r_j) \sim \hat{T}(s_j, a_j) \).
      
      Add \((s_j, a_j, r_j, s_{j+1})\) to \( D_{\text{sample}} \).
    }
  }
  
  \For{\( k = 1 \) to \( K \)}{
    Sample a batch \( \hat{D}_{\text{expert}} \subset D_{\text{expert}} \).
    
    Sample a batch \( \hat{D}_{\text{sample}} \subset D_{\text{sample}} \).
    
    Optimize \( r_{\psi} \) according to \eqref{eq8} using \( \hat{D}_{\text{sample}} \cup \hat{D}_{\text{expert}} \)
  }
   
  Conservatively evaluate \( \pi_{\phi} \) by solving \eqref{eq5} to update \(\hat{Q}_{\phi}\) \ using samples from \( {D}_{\text{sample}} \cup {D}_{\text{expert}} \).
  
  Improve policy under state marginal of \( d_{\lambda} \) by solving \eqref{eq6} to update \( \pi_{\phi} \).
}
\end{algorithm}

\section{Experimental Setup}
We validated our proposed approach in a heparin dosing environment. This section describes the dataset and the implementation of the treatment model.

\subsection{Dataset}
We utilized the MIMIC-III database, which contains comprehensive treatment data for over 40,000 patients. The dataset was collected from an intensive care unit (ICU) and includes more than 60 patient variables, such as vital signs, demographic information, health scores, and prescription details. We selected patients who received heparin treatment, resulting in a dataset comprising data from 1,911 individuals extracted from MIMIC-III. Of this dataset, 80\% (n = 1,528) was allocated for training, and 20\% (n = 383) was reserved for testing. The preprocessing steps are as follows:

We collected health information, anticoagulation-related laboratory test results, vital signs, and outcomes of heparin administration for patients aged 18 and older. Data were extracted for a duration ranging from a minimum of 7 hours to a maximum of 72 hours from the time of heparin administration. Records with durations shorter than 7 hours were excluded due to insufficient data. One of the main heparin injection methods is continuous intravenous infusion (drip). To closely monitor temporal changes, we segmented the data into 1-hour intervals.

Medical data often contain missing values due to factors such as prolonged measurement intervals, human omission, or unmeasurable parameters. To address missing values, we first calculated the missing rate for each feature and excluded those with high missing rates from our analysis. Next, we removed outliers using frequency histograms and Tukey’s method, and imputed missing values with values derived from other features. For data that remained missing but could be inferred from other variables, such as SOFA and GCS scores, the missing values were calculated based on the available measurements. Subsequently, the remaining missing values were imputed using the sample-and-hold method. Finally, we employed data-driven approaches, including K-Nearest Neighbors (KNN) and linear imputation methods, to handle any missing values not addressed by the previous imputation techniques. After completing the data imputation, we removed medically nonsensical outliers and performed normalization.

The final variables included age, gender, Glasgow Coma Scale (GCS), diastolic and systolic arterial blood pressure (DBP and SBP), respiratory rate (RR), hemoglobin (HGB), temperature, white blood cell count (WBC), platelet count, activated partial thromboplastin time (aPTT), prothrombin time (PT), arterial carbon dioxide (ACD), creatinine, bilirubin, international normalized ratio of prothrombin (INR), and weight.

\subsection{Practical Implementation}
To conduct RL training, we first established the MDP problem. State represents the current condition of the patient. We derived the state using all indicators from the preprocessing results except for aPTT and heparin dosage. Action corresponds to the clinicians’ heparin treatment records. We categorized heparin IV injection levels into six distinct classes for the action space based on quantiles. State and action dimensions are 16 and 6, respectively. Next state denotes the patient’s condition one hour after the current state. Additionally, we defined a pre-determined reward function ($r_p$) to verify whether the policy influenced by $r_\psi$ aligns with clinical outcomes. The aPTT is a critical indicator of blood clotting ability. Following methodologies from prior research \cite{14}, $r_p$ assigns rewards close to 1 for aPTT values within the therapeutic range (60–100 sec) and approximately -1 for values outside this range. The formula used is $r_p = \frac{2}{1+e^{-(\text{aPTT}-60)}}-\frac{2}{1+e^{-(\text{aPTT}-100)}}-1$.

The core component of OMG-RL is the acquisition of expert trajectories. Given that the MIMIC-III dataset comprises actual medication records from clinical specialists, it is sufficiently justified to assume that these trajectories exhibit at least sub-optimal characteristics inherent to expert behavior. The dynamic model was trained using the MIMIC-III training set, applying the $r_p$ function. To construct an ensemble model, we trained seven neural networks and selected the five with the highest performance metrics. The dynamic model was developed using four fully connected layers. The $\lambda$ parameter for updating the Q-network was fixed at 0.5 to ensure that both the sample distribution and the expert distribution were equally represented. The reinforcement learning algorithms employed to implement COMBO and OMG-RL approaches utilized the Soft Actor-Critic (SAC) \cite{48}, wherein both the actor and critic comprised three fully connected layers. Additionally, the reward network consisted of three fully connected layers.

\section{Results}
\subsection{Dynamic Model Performance}
\begin{figure}[h]
    \centering
    \includegraphics[width=0.65\linewidth]{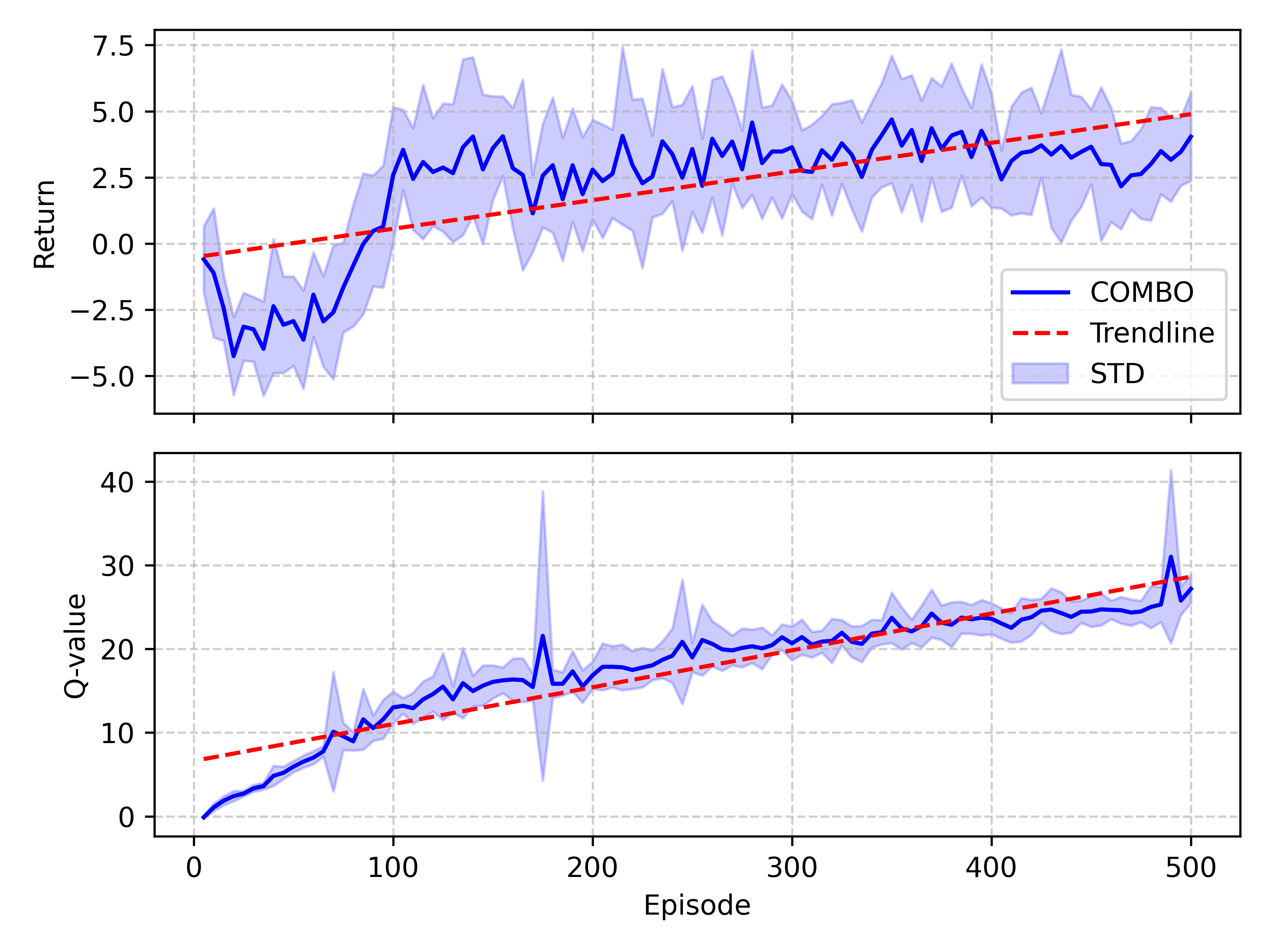}
    \caption{Changes in Return ($r_p$) and Q-Value of COMBO.}
    \label{fig: figure2}
\end{figure}

We implemented the COMBO algorithm to evaluate whether the dynamic model was adequately trained. Utilizing the clinical dataset and the dynamic model, we trained COMBO over 500 episodes and calculated the mean and standard deviation of test returns and test Q-values across a total of ten experiments. The test results are presented in Fig. \ref{fig: figure2}. The returns increased with the number of episodes, exhibiting a tendency to converge at an average value of approximately 4. Similarly, the Q-values continuously increased. These findings confirm that exploration and exploitation via the dynamic model positively influence the reinforcement of the agent's policy.

\begin{figure}[h!]
    \centering
    \begin{subfigure}{0.46\textwidth}
        \includegraphics[width=\linewidth]{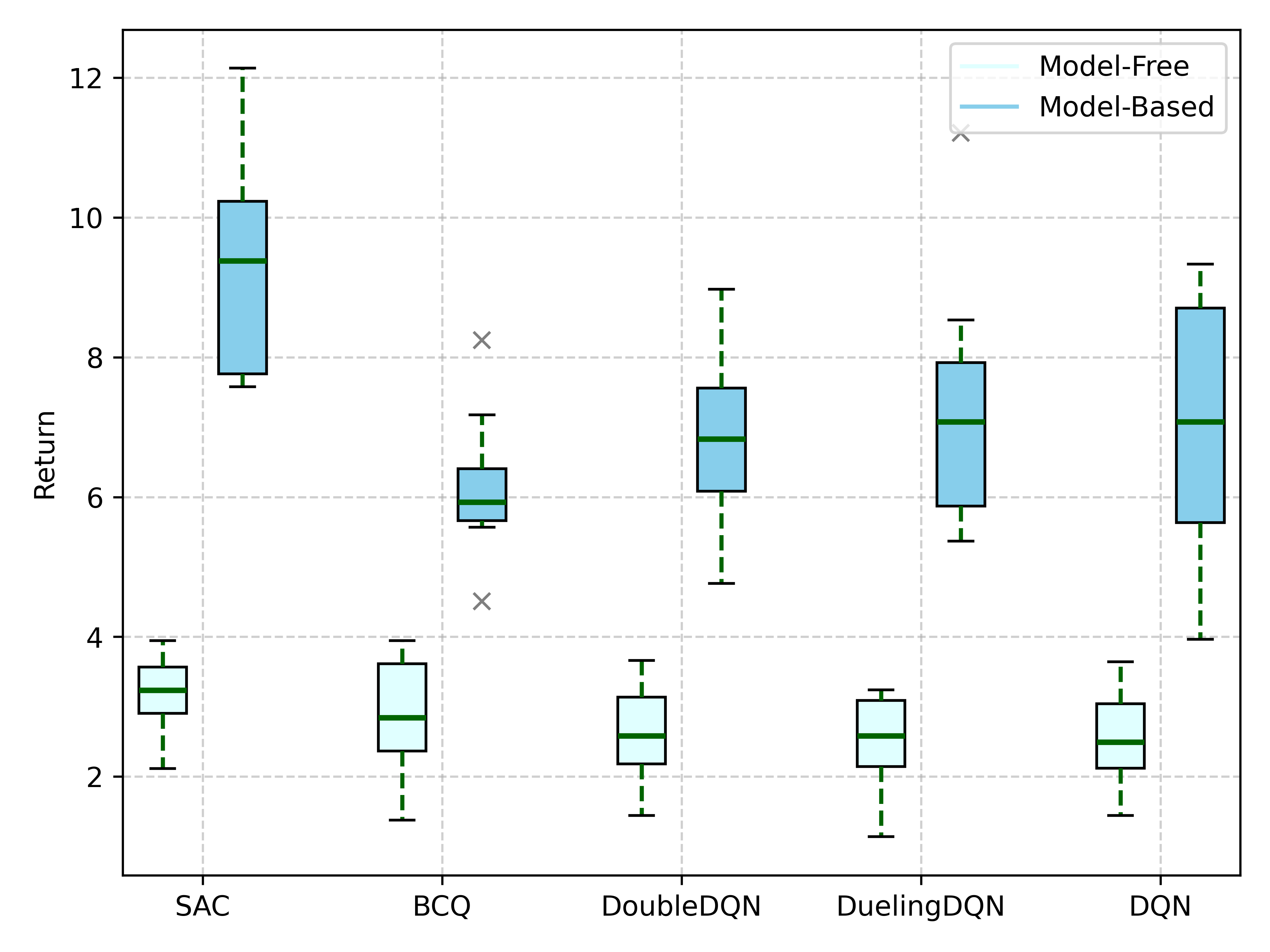}
        \caption{Comparison of returns ($r_{p}$).}
        \label{fig: figure3}
    \end{subfigure}
    \hspace{0.02\textwidth}
    \begin{subfigure}{0.46\textwidth}
        \includegraphics[width=\linewidth]{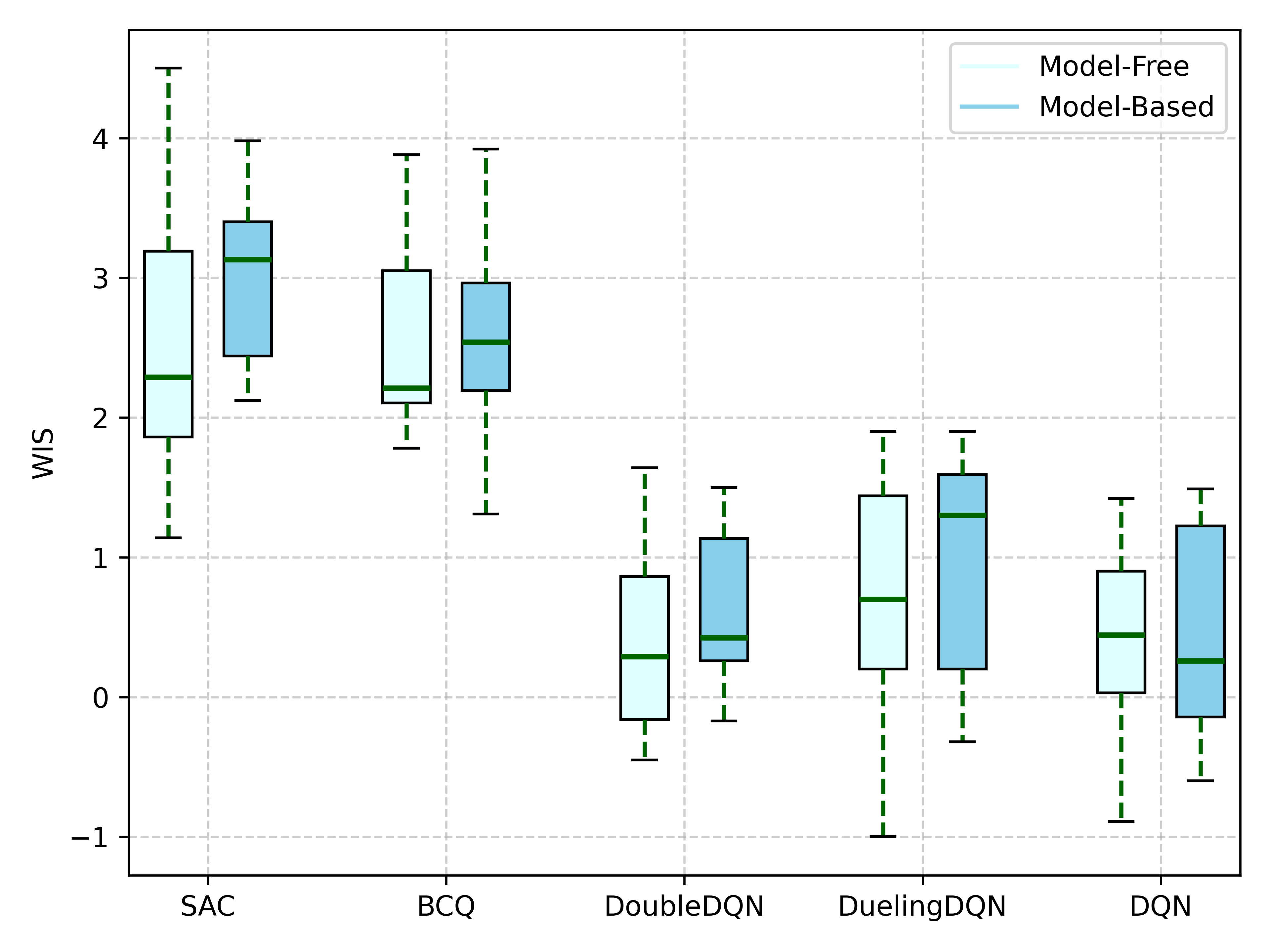}
        \caption{Comparison of returns (WIS).}
        \label{fig: figure4}
    \end{subfigure}
    \caption{Comparison of returns between model-based and model-free approaches: (a) returns ($r_{p}$) and (b) returns (WIS).}
    \label{fig:comparison_returns}
\end{figure}

Additionally, to assess the generalizability of the dynamic model, we conducted a comparative evaluation of the SAC, BCQ, Double DQN, Dueling DQN, and DQN algorithms. Each algorithm was trained using both model-free and Dyna-style model-based approaches and subsequently tested with the dynamic model. We performed five experiments in total and visualized the obtained test return results using box plots. The test results are shown in Fig. \ref{fig: figure3}. The returns for SAC trained in a model-free manner ranged between 2 and 4, whereas those trained in a model-based manner ranged between 8 and 10. As is commonly known, introducing a model universally enhanced the policy performance across all algorithms.

Furthermore, to account for potential biases that may arise during evaluations using the dynamic model, we employed the weighted importance sampling (WIS) technique, an off-policy evaluation method, to assess each algorithm under identical settings. The tests utilized MIMIC-III test set. The results are illustrated in Fig. \ref{fig: figure4}. For SAC, DoubleDQN, DuelingDQN, and DQN, the WIS values for the model-based approaches were approximately 0.2, 0.25, 0.1, and 0.3 higher, respectively, than those of the model-free approach. Overall, the model-based approach demonstrated superior performance; however, the magnitude of this improvement was smaller compared to the results shown in Fig. \ref{fig: figure3} because the total reward sum of the test trajectories was fixed. Consequently, even with the application of superior policies, deviations within a specific reward range were constrained. Additional evaluations using other off-policy evaluation (OPE) techniques are required. Nonetheless, depending on the evaluation objectives, the results from Figs. \ref{fig: figure2} and \ref{fig:comparison_returns} confirm that the dynamic model was appropriately trained.

\subsection{OMG-RL Performance}
\begin{figure}[h]
    \centering
    \includegraphics[width=0.65\linewidth]{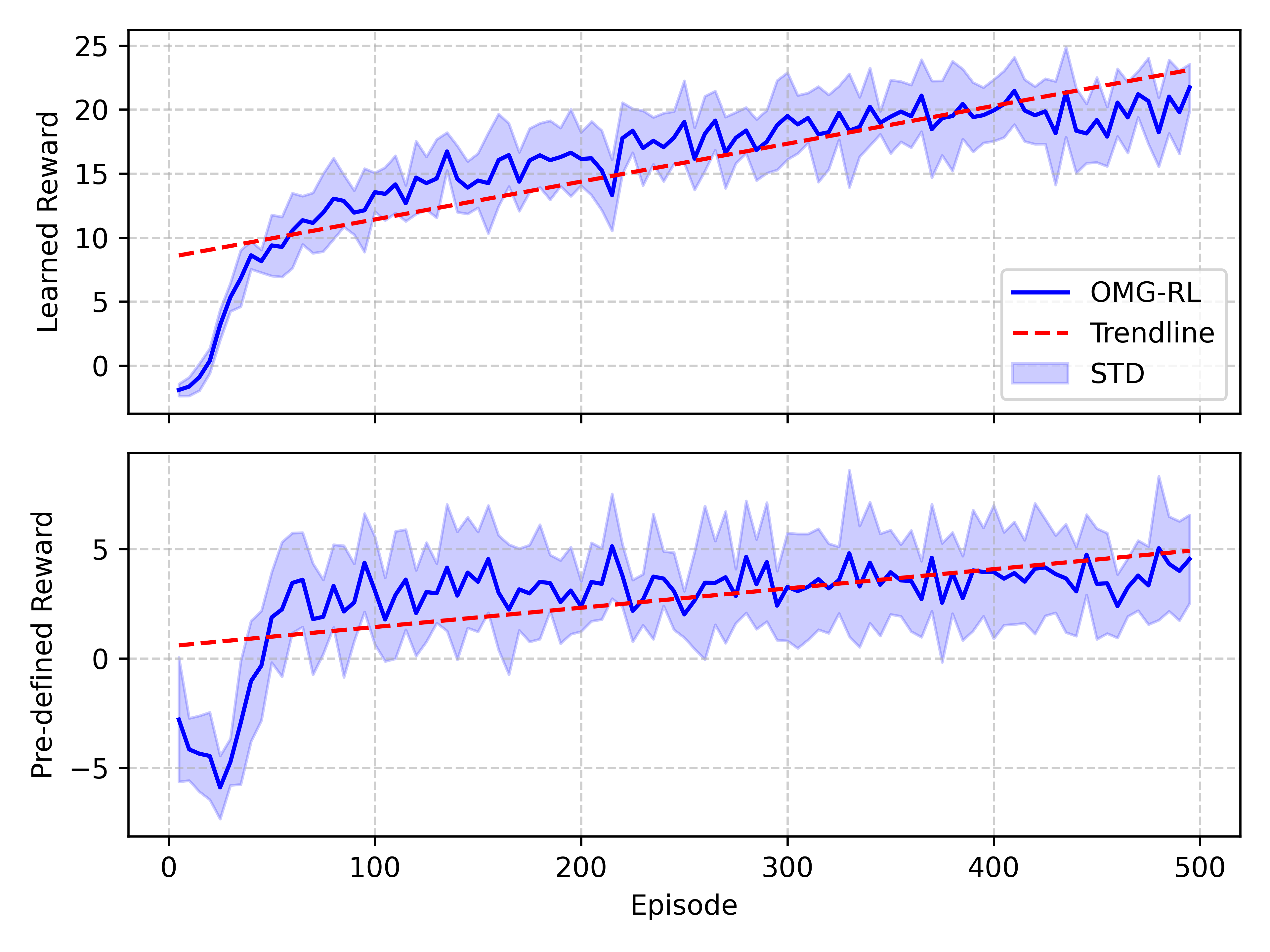}
    \caption{Changes in $r_\psi$ and $r_p$ of OMG-RL.}
    \label{fig: figure5}
\end{figure}

We evaluated the learning performance of OMG-RL using both the learned reward ($r_\psi$) derived from the reward network and the pre-defined reward ($r_p$) obtainable from the dynamic model. The purpose of evaluating $r_\psi$ was to determine whether the model was learning in a manner that enhances rewards, while $r_p$ was used to assess the tendencies of the learned policy with respect to key anticoagulation indicators. Training and evaluation were conducted over 500 episodes, and we calculated the mean and standard deviation of test $r_\psi$ and test $r_p$ across a total of ten experiments. The test results are presented in Fig. \ref{fig: figure5}. The learned reward $r_\psi$ increased with the number of episodes, showing a tendency to converge to an average value of approximately 20. Similarly, $r_p$ also exhibited an increasing trend, converging to an average value of about 5. These findings confirm that the OMG-RL framework is capable of addressing the offline IRL problem and that the data-driven reward function appropriately reflects the properties of key medical indicators.

\begin{figure}[h]
    \centering
    \includegraphics[width=0.6\linewidth]{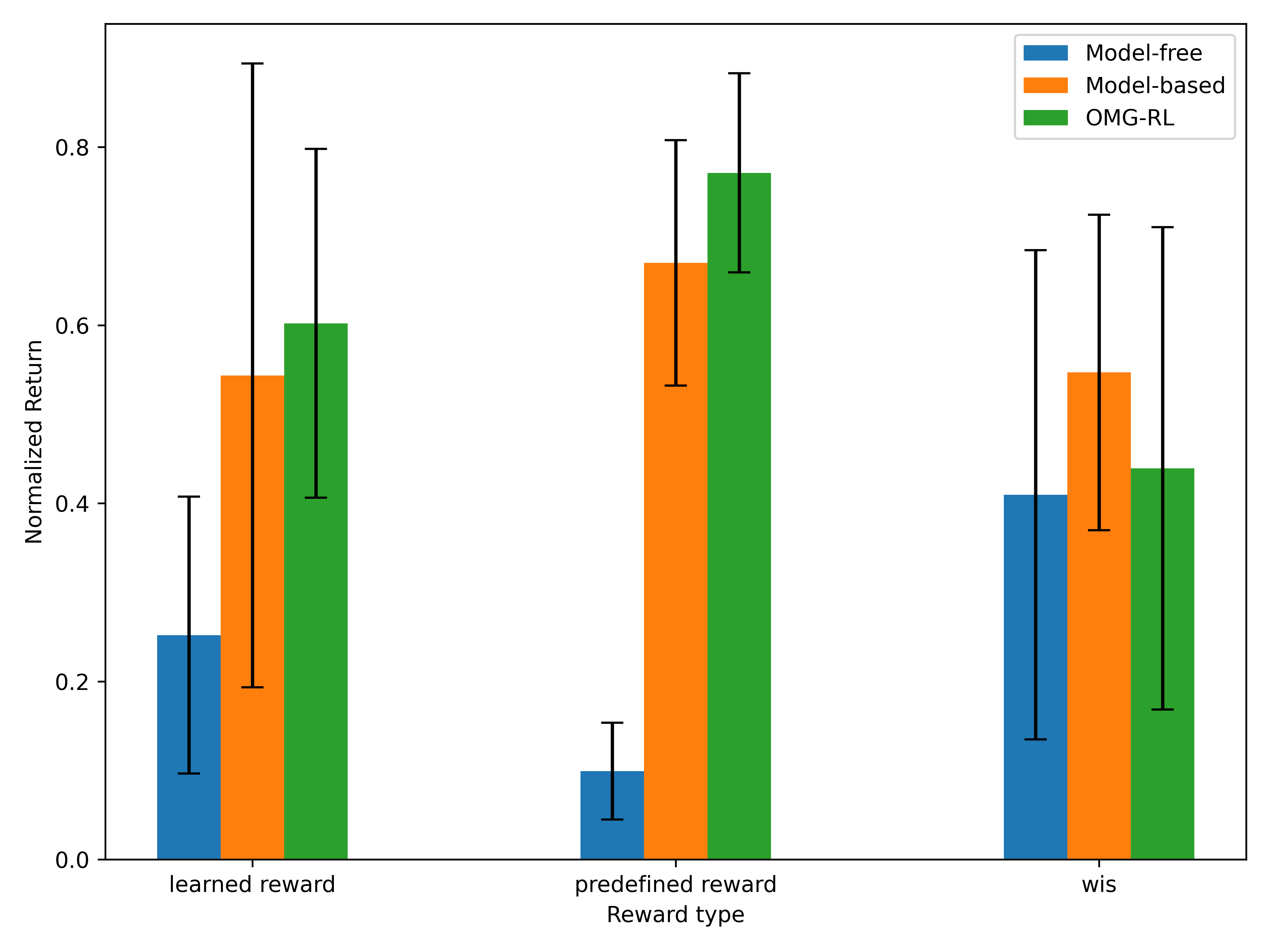}
    \caption{Comparison of normalized returns across three reward types ($r_\psi$, $r_p$, WIS) for OMG-RL, model-based, and model-free approaches.}
    \label{fig: figure6}
\end{figure}

Subsequently, we compared the policies of OMG-RL with model-based and model-free approaches and evaluated whether the learned reward function is medically meaningful using three evaluation metrics. Training and evaluation were conducted over 500 episodes, and the normalized means and standard deviations of five experiments were visualized using bar graphs. The results are shown in Fig. \ref{fig: figure6}. The performance of $r_\psi$ and $r_p$ ranked highest in the order of model-free, model-based, and OMG-RL approaches. Evaluations using WIS indicated that the model-based methodologies achieved the highest performance. Furthermore, the model-based policy trained with $r_p$ also demonstrated significant performance on $r_\psi$, suggesting that the policy trained on aPTT may be sub-optimal in $r_\psi$. The WIS results showed minimal performance differences among the three algorithms, consistent with the analysis of the results in Fig. \ref{fig: figure4}.

\subsection{Model Anaysis}
\begin{figure}[h]
    \centering
    \includegraphics[width=0.6\linewidth]{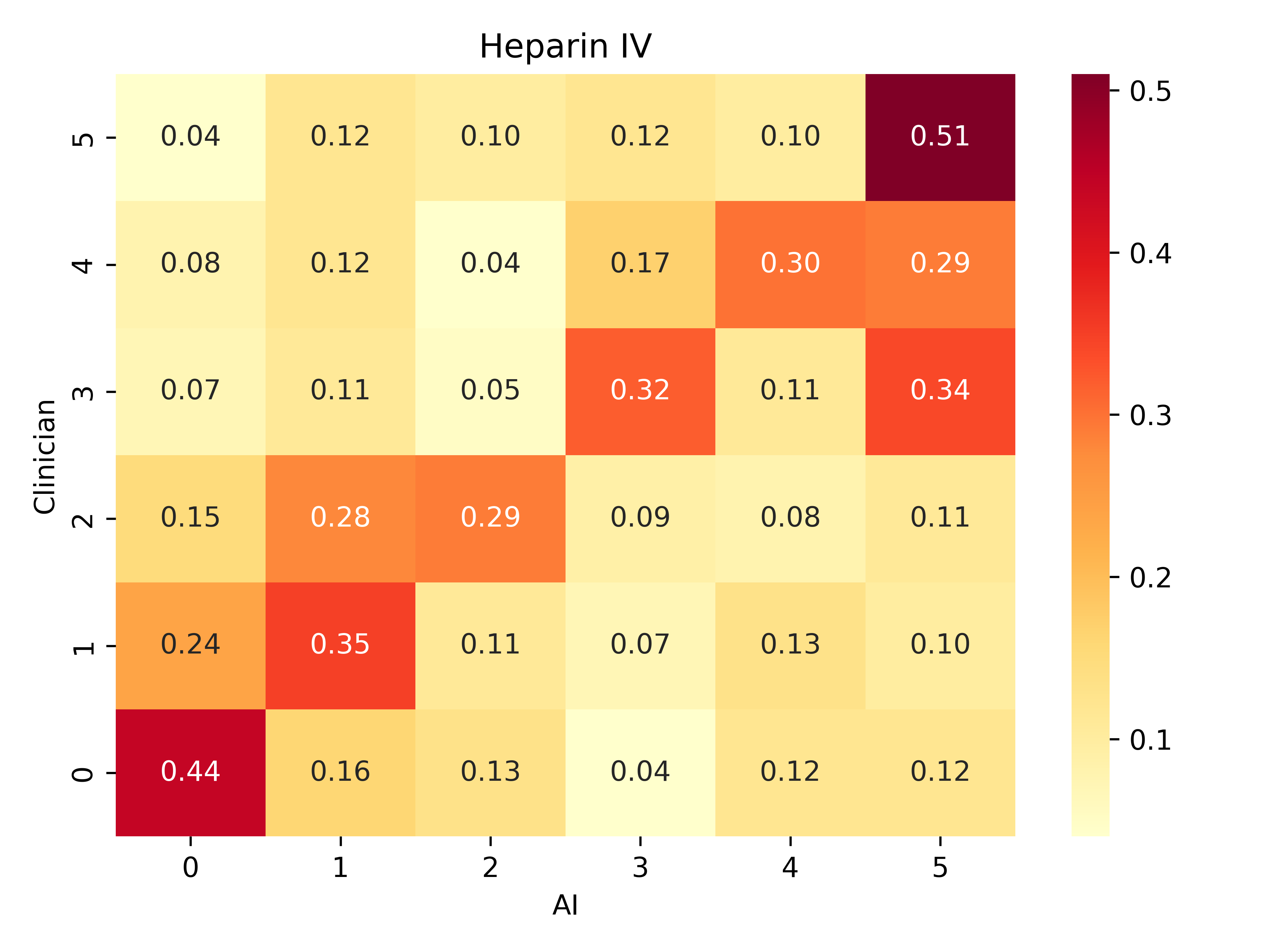}
    \caption{Agreement between OMG-RL and clinicians' treatment decisions.}
    \label{fig: figure7}
\end{figure}

To investigate how the proposed OMG-RL model modifies therapeutic behaviors to enhance treatment outcomes, heatmap graphs were employed. Fig. \ref{fig: figure7} illustrates the results of individual treatment analyses based on Heparin dosages. The activated lines along the diagonal indicate a high degree of concordance between the individual treatment decisions of the AI model and those of clinicians under identical patient conditions. Predominantly, high probabilities were observed around the diagonal, suggesting that AI and clinicians generally share similar prescribing strategies. Notably, prescriptions at Stage 1 and Stage 6 exhibited relatively high concordance rates of 44\% and 51\%, respectively. Additionally, higher probabilities were observed in the lower-left and upper-right corners, implying that the AI proposes more extreme treatment strategies compared to the conservative approaches typically employed in clinical practice.

\begin{figure}[h]
    \centering
    \includegraphics[width=0.8\linewidth]{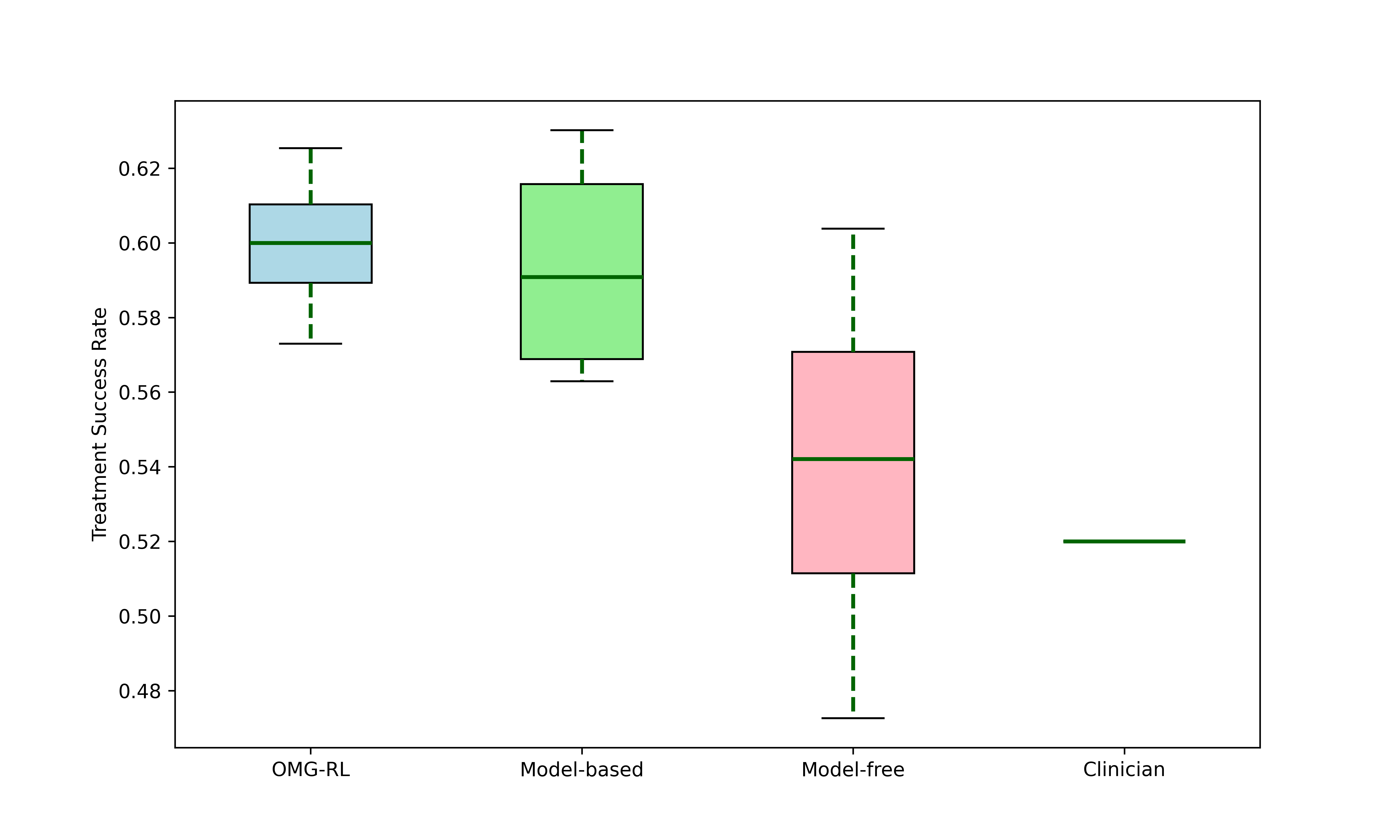}
    \caption{Comparison of treatment success rates among OMG-RL, model-based, model-free, and clinicians.}
    \label{fig: figure8}
\end{figure}

The effectiveness of the OMG-RL model in increasing the success rate of heparin therapy was evaluated. Treatment segments within individual trajectories were deemed successful if the reward persisted above 0.8 for at least two hours; otherwise, they were considered failures. For clinicians, evaluation was conducted using test trajectories from the MIMIC-III dataset. Other algorithms were assessed by rolling out a dynamic model over 36 steps (36 hours), considering the treatment successful if a successful segment appeared during this period. A total of 380 episodes were conducted. Both the number of episodes and steps were selected based on the statistics of the MIMIC-III dataset. The results of ten experimental runs were visualized using box plots (Figure \ref{fig: figure8}). The treatment success rate of OMG-RL ranged from approximately 0.59 to 0.61, with a median around 0.60. The model-based approach demonstrated performance comparable to OMG-RL. In contrast, the model-free approach exhibited treatment success rates ranging from approximately 0.50 to 0.57, with a median of about 0.54. Both OMG-RL and the model-based approach showed similar performance, achieving high treatment success rates, thereby indicating the effectiveness of both methodologies. Overall, the AI-based approaches exhibited higher treatment success rates compared to clinicians, suggesting that the AI model has the potential to deliver improved therapeutic performance over existing clinical treatment strategies.

\begin{figure}[h!]
    \centering
    \begin{subfigure}{0.47\textwidth}
        \centering
        \includegraphics[width=\linewidth]{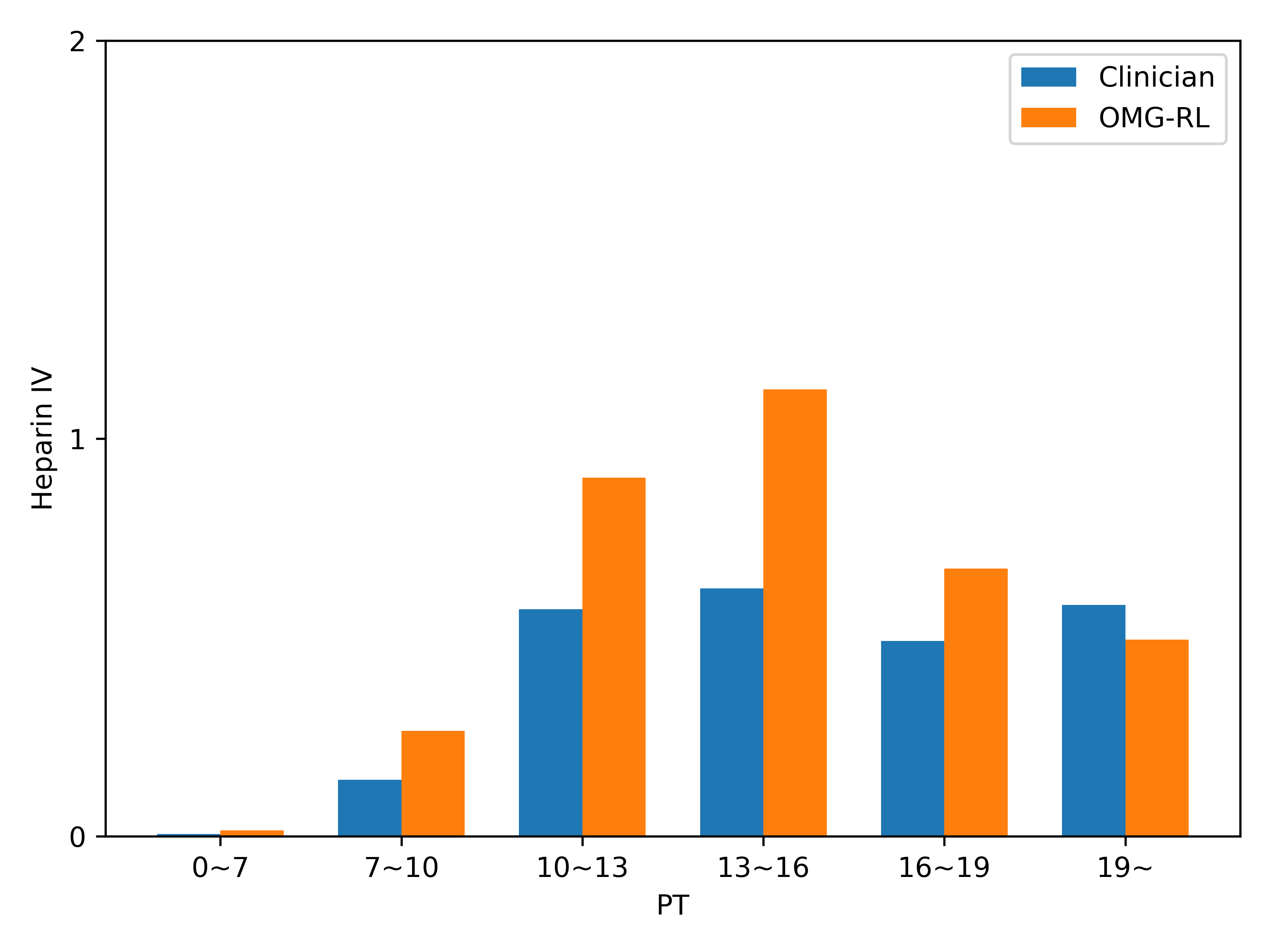}
        \caption{Treatment tendencies for PT.}
        \label{fig: figure9}
    \end{subfigure}
    \hspace{0.01\textwidth}
    \begin{subfigure}{0.47\textwidth}
        \centering
        \includegraphics[width=\linewidth]{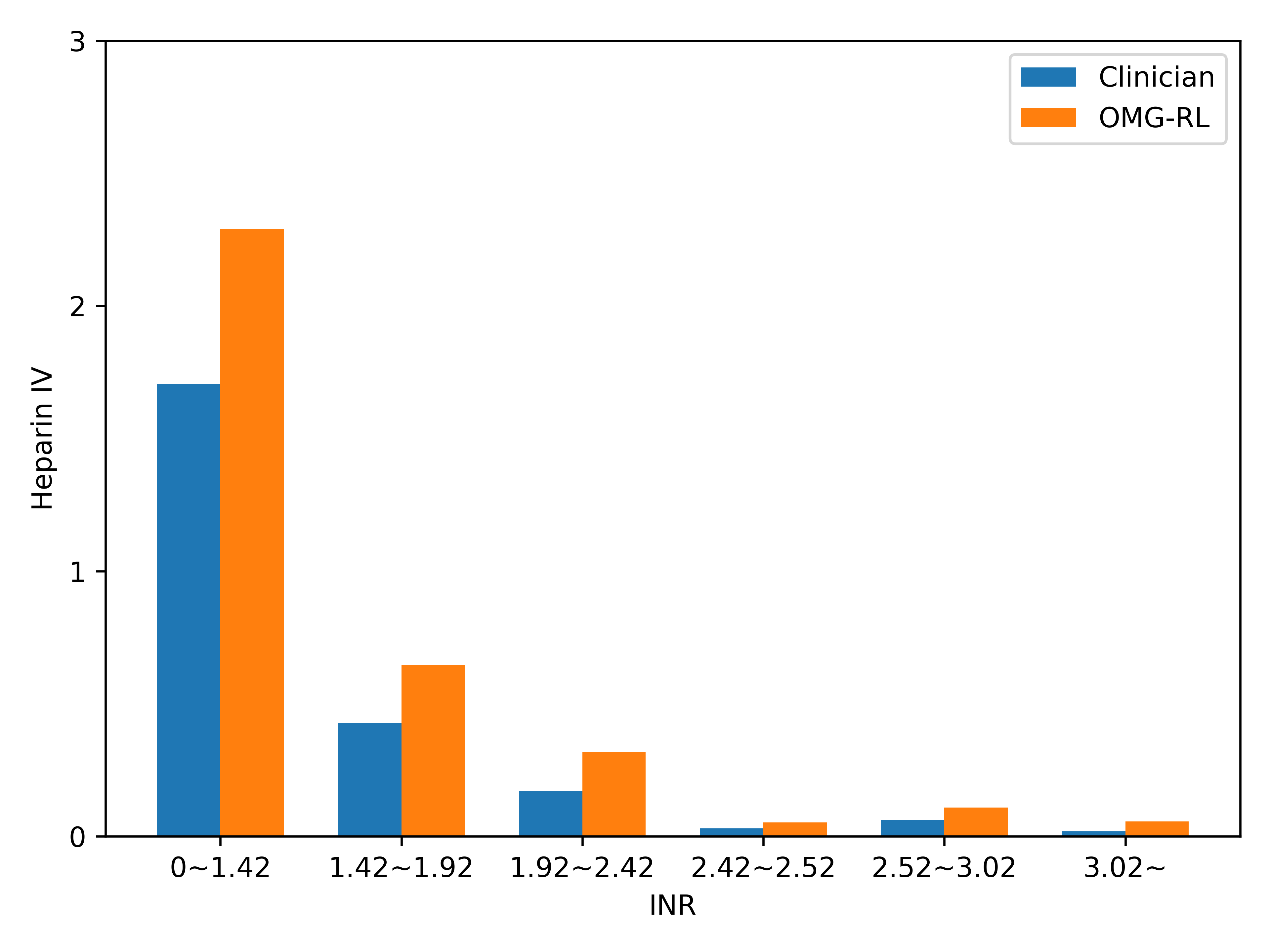}
        \caption{Treatment tendencies for INR.}
        \label{fig: figure10}
    \end{subfigure}
    \caption{OMG-RL and clinicians' treatment tendencies for prothrombin time(PT) and international normalized ratio of prothrombin(INR) values.}
    \label{fig:combined_figures2}
\end{figure}

We evaluated the administration tendencies of clinicians and the proposed model based on status indicators. PT and INR serve as metrics for assessing anticoagulation. We examined the average medication dosages corresponding to each state value, with the respective results illustrated in Fig. \ref{fig:combined_figures2}. The OMG-RL model concentrates on administering heparin IV within the PT range of 10–16, potentially reflecting an aggressive treatment strategy. In contrast, clinicians demonstrate a more balanced administration tendency compared to OMG-RL, adopting a relatively conservative strategy that does not strongly favor a specific PT range. A key observation is that the administration actions are included between 0 and 1, indicating that the actual dosages are minimal. These administration tendencies are derived from patient records. The normal PT range for adults is 11–15 seconds. Considering these factors, OMG-RL's strategy appears to be partially rational. This strategy involves incrementing dosages as PT approaches the normal range to facilitate convergence towards normality and reducing dosages when PT deviates from the normal range. As depicted in Fig. \ref{fig: figure10}, OMG-RL tends to administer heparin IV more aggressively than clinicians when INR values are low. This suggests that the AI model adopts an aggressive strategy, anticipating higher therapeutic efficacy when INR values are low. Conversely, clinicians employ a relatively conservative approach when INR values are low and display administration patterns similar to OMG-RL as INR increases. In higher INR ranges, both AI and clinicians adopt a conservative strategy by significantly reducing administration frequency. Generally, considering that the INR range for anticoagulated patients is 2.0 to 3.0, the administration trend appears to be very reasonable.

\section{Conclusion}

In this study, we have departed from conventional methods that typically rely on clinical knowledge to define reward functions. Instead, we implemented IRL, which derives reward estimations directly from data, to inform the learning of RL policies. Our data source, the MIMIC-III electronic medical records, served as a foundation for this analysis, with specific attention to preprocessing heparin administration records for the purpose of model validation. This led to the development of the Offline Model-based Guided Reward Learning (OMG-RL) model, designed to facilitate IRL within an offline framework.

Throughout various experiments, our approach demonstrated significant validity. Particularly, we noted that the behavior of \(r_{\psi}\) closely mirrored \(r_p\), established through clinical insights, underscoring the effectiveness of IRL in capturing complex decision-making processes inherent in medical treatment. This finding supports the potential of IRL-based policy learning to reach substantial performance benchmarks, comparable to those grounded in traditional domain expertise. We anticipate that our methodology will offer a robust alternative for managing medication dosing tasks and potentially influence other areas where reward definition poses a substantial challenge.
Additionally, we confirmed through various model analysis techniques that the AI's administration tendencies are highly rational and capable of achieving a high success rate in heparin therapy.

However, the study is not without limitations. The discrete nature of the action space used in our experiments stands in contrast to the continuous variables typically encountered in medication dosing, suggesting a need for models that accommodate continuous action frameworks to enhance predictive accuracy. Furthermore, while our theoretical approach has been substantiated through methodological rigor, additional empirical validation is necessary across broader demographic and geographic patient data sets to ensure generalizability and applicability in real-world clinical settings. Future research will aim to refine our reinforcement learning framework to address these challenges, enhancing its practical relevance and efficacy.

\bibliographystyle{cas-model2-names}
\bibliography{manuscript}

\begin{thebibliography}{41}
\expandafter\ifx\csname natexlab\endcsname\relax\def\natexlab#1{#1}\fi
\providecommand{\url}[1]{\texttt{#1}}
\providecommand{\href}[2]{#2}
\providecommand{\path}[1]{#1}
\providecommand{\DOIprefix}{doi:}
\providecommand{\ArXivprefix}{arXiv:}
\providecommand{\URLprefix}{URL: }
\providecommand{\Pubmedprefix}{pmid:}
\providecommand{\doi}[1]{\href{http://dx.doi.org/#1}{\path{#1}}}
\providecommand{\Pubmed}[1]{\href{pmid:#1}{\path{#1}}}
\providecommand{\bibinfo}[2]{#2}
\ifx\xfnm\relax \def\xfnm[#1]{\unskip,\space#1}\fi
\bibitem[{Abbeel and Ng(2004)}]{17}
\bibinfo{author}{Abbeel, P.}, \bibinfo{author}{Ng, A.Y.}, \bibinfo{year}{2004}.
\newblock \bibinfo{title}{Apprenticeship learning via inverse reinforcement learning}, in: \bibinfo{booktitle}{Proceedings of the twenty-first international conference on Machine learning}, p.~\bibinfo{pages}{1}.
\bibitem[{Abdulhai et~al.(2022)Abdulhai, Jaques and Levine}]{37}
\bibinfo{author}{Abdulhai, M.}, \bibinfo{author}{Jaques, N.}, \bibinfo{author}{Levine, S.}, \bibinfo{year}{2022}.
\newblock \bibinfo{title}{Basis for intentions: Efficient inverse reinforcement learning using past experience}.
\newblock \bibinfo{journal}{arXiv preprint arXiv:2208.04919} .
\bibitem[{Arora and Doshi(2021)}]{18}
\bibinfo{author}{Arora, S.}, \bibinfo{author}{Doshi, P.}, \bibinfo{year}{2021}.
\newblock \bibinfo{title}{A survey of inverse reinforcement learning: Challenges, methods and progress}.
\newblock \bibinfo{journal}{Artificial Intelligence} \bibinfo{volume}{297}, \bibinfo{pages}{103500}.
\bibitem[{Baucum et~al.(2020)Baucum, Khojandi and Vasudevan}]{42}
\bibinfo{author}{Baucum, M.}, \bibinfo{author}{Khojandi, A.}, \bibinfo{author}{Vasudevan, R.}, \bibinfo{year}{2020}.
\newblock \bibinfo{title}{Improving deep reinforcement learning with transitional variational autoencoders: A healthcare application}.
\newblock \bibinfo{journal}{IEEE Journal of Biomedical and Health Informatics} \bibinfo{volume}{25}, \bibinfo{pages}{2273--2280}.
\bibitem[{Baucum et~al.(2022)Baucum, Khojandi, Vasudevan and Davis}]{43}
\bibinfo{author}{Baucum, M.}, \bibinfo{author}{Khojandi, A.}, \bibinfo{author}{Vasudevan, R.}, \bibinfo{author}{Davis, R.}, \bibinfo{year}{2022}.
\newblock \bibinfo{title}{Adapting reinforcement learning treatment policies using limited data to personalize critical care}.
\newblock \bibinfo{journal}{INFORMS Journal on Data Science} \bibinfo{volume}{1}, \bibinfo{pages}{27--49}.
\bibitem[{Breslin et~al.(2004)Breslin, Mirakhur, Reid and Kyle}]{5}
\bibinfo{author}{Breslin, D.}, \bibinfo{author}{Mirakhur, R.}, \bibinfo{author}{Reid, J.}, \bibinfo{author}{Kyle, A.}, \bibinfo{year}{2004}.
\newblock \bibinfo{title}{Manual versus target-controlled infusions of propofol}.
\newblock \bibinfo{journal}{Anaesthesia} \bibinfo{volume}{59}, \bibinfo{pages}{1059--1063}.
\bibitem[{Ebert et~al.(2018)Ebert, Finn, Dasari, Xie, Lee and Levine}]{27}
\bibinfo{author}{Ebert, F.}, \bibinfo{author}{Finn, C.}, \bibinfo{author}{Dasari, S.}, \bibinfo{author}{Xie, A.}, \bibinfo{author}{Lee, A.}, \bibinfo{author}{Levine, S.}, \bibinfo{year}{2018}.
\newblock \bibinfo{title}{Visual foresight: Model-based deep reinforcement learning for vision-based robotic control}.
\newblock \bibinfo{journal}{arXiv preprint arXiv:1812.00568} .
\bibitem[{Finn et~al.(2016)Finn, Levine and Abbeel}]{33}
\bibinfo{author}{Finn, C.}, \bibinfo{author}{Levine, S.}, \bibinfo{author}{Abbeel, P.}, \bibinfo{year}{2016}.
\newblock \bibinfo{title}{Guided cost learning: Deep inverse optimal control via policy optimization}, in: \bibinfo{booktitle}{International conference on machine learning}, \bibinfo{organization}{PMLR}. pp. \bibinfo{pages}{49--58}.
\bibitem[{Garg et~al.(2021)Garg, Chakraborty, Cundy, Song and Ermon}]{36}
\bibinfo{author}{Garg, D.}, \bibinfo{author}{Chakraborty, S.}, \bibinfo{author}{Cundy, C.}, \bibinfo{author}{Song, J.}, \bibinfo{author}{Ermon, S.}, \bibinfo{year}{2021}.
\newblock \bibinfo{title}{Iq-learn: Inverse soft-q learning for imitation}.
\newblock \bibinfo{journal}{Advances in Neural Information Processing Systems} \bibinfo{volume}{34}, \bibinfo{pages}{4028--4039}.
\bibitem[{Ghoneim and Weiskopf(2000)}]{6}
\bibinfo{author}{Ghoneim, M.M.}, \bibinfo{author}{Weiskopf, R.B.}, \bibinfo{year}{2000}.
\newblock \bibinfo{title}{Awareness during anesthesia}.
\newblock \bibinfo{journal}{The Journal of the American Society of Anesthesiologists} \bibinfo{volume}{92}, \bibinfo{pages}{597--597}.
\bibitem[{Gill et~al.(2016)Gill, Machavaram, Rose and Chetty}]{2}
\bibinfo{author}{Gill, K.L.}, \bibinfo{author}{Machavaram, K.K.}, \bibinfo{author}{Rose, R.H.}, \bibinfo{author}{Chetty, M.}, \bibinfo{year}{2016}.
\newblock \bibinfo{title}{Potential sources of inter-subject variability in monoclonal antibody pharmacokinetics}.
\newblock \bibinfo{journal}{Clinical pharmacokinetics} \bibinfo{volume}{55}, \bibinfo{pages}{789--805}.
\bibitem[{Haarnoja et~al.(2018)Haarnoja, Zhou, Abbeel and Levine}]{48}
\bibinfo{author}{Haarnoja, T.}, \bibinfo{author}{Zhou, A.}, \bibinfo{author}{Abbeel, P.}, \bibinfo{author}{Levine, S.}, \bibinfo{year}{2018}.
\newblock \bibinfo{title}{Soft actor-critic: Off-policy maximum entropy deep reinforcement learning with a stochastic actor}, in: \bibinfo{booktitle}{International conference on machine learning}, \bibinfo{organization}{PMLR}. pp. \bibinfo{pages}{1861--1870}.
\bibitem[{Hirsh et~al.(2001)Hirsh, Warkentin, Shaughnessy, Anand, Halperin, Raschke, Granger, Ohman and Dalen}]{4}
\bibinfo{author}{Hirsh, J.}, \bibinfo{author}{Warkentin, T.E.}, \bibinfo{author}{Shaughnessy, S.G.}, \bibinfo{author}{Anand, S.S.}, \bibinfo{author}{Halperin, J.L.}, \bibinfo{author}{Raschke, R.}, \bibinfo{author}{Granger, C.}, \bibinfo{author}{Ohman, E.M.}, \bibinfo{author}{Dalen, J.E.}, \bibinfo{year}{2001}.
\newblock \bibinfo{title}{Heparin and low-molecular-weight heparin mechanisms of action, pharmacokinetics, dosing, monitoring, efficacy, and safety}.
\newblock \bibinfo{journal}{Chest} \bibinfo{volume}{119}, \bibinfo{pages}{64S--94S}.
\bibitem[{Jack(2001)}]{3}
\bibinfo{author}{Jack, H.}, \bibinfo{year}{2001}.
\newblock \bibinfo{title}{Guide to anticoagulant therapy: heparin}.
\newblock \bibinfo{journal}{Circulation} \bibinfo{volume}{103}, \bibinfo{pages}{2994--3018}.
\bibitem[{Jia et~al.(2020)Jia, Burden, Lawton and Habli}]{10}
\bibinfo{author}{Jia, Y.}, \bibinfo{author}{Burden, J.}, \bibinfo{author}{Lawton, T.}, \bibinfo{author}{Habli, I.}, \bibinfo{year}{2020}.
\newblock \bibinfo{title}{Safe reinforcement learning for sepsis treatment}, in: \bibinfo{booktitle}{2020 IEEE International Conference on Healthcare Informatics (ICHI)}, \bibinfo{organization}{IEEE}. pp. \bibinfo{pages}{1--7}.
\bibitem[{Johnson et~al.(2016)Johnson, Pollard, Shen, Lehman, Feng, Ghassemi, Moody, Szolovits, Anthony~Celi and Mark}]{47}
\bibinfo{author}{Johnson, A.E.}, \bibinfo{author}{Pollard, T.J.}, \bibinfo{author}{Shen, L.}, \bibinfo{author}{Lehman, L.w.H.}, \bibinfo{author}{Feng, M.}, \bibinfo{author}{Ghassemi, M.}, \bibinfo{author}{Moody, B.}, \bibinfo{author}{Szolovits, P.}, \bibinfo{author}{Anthony~Celi, L.}, \bibinfo{author}{Mark, R.G.}, \bibinfo{year}{2016}.
\newblock \bibinfo{title}{Mimic-iii, a freely accessible critical care database}.
\newblock \bibinfo{journal}{Scientific data} \bibinfo{volume}{3}, \bibinfo{pages}{1--9}.
\bibitem[{Klein et~al.(2011)Klein, Geist and Pietquin}]{35}
\bibinfo{author}{Klein, E.}, \bibinfo{author}{Geist, M.}, \bibinfo{author}{Pietquin, O.}, \bibinfo{year}{2011}.
\newblock \bibinfo{title}{Batch, off-policy and model-free apprenticeship learning}, in: \bibinfo{booktitle}{European Workshop on Reinforcement Learning}, \bibinfo{organization}{Springer}. pp. \bibinfo{pages}{285--296}.
\bibitem[{Lazzati et~al.(2024)Lazzati, Mutti and Metelli}]{52}
\bibinfo{author}{Lazzati, F.}, \bibinfo{author}{Mutti, M.}, \bibinfo{author}{Metelli, A.M.}, \bibinfo{year}{2024}.
\newblock \bibinfo{title}{Offline inverse rl: New solution concepts and provably efficient algorithms}.
\newblock \bibinfo{journal}{arXiv preprint arXiv:2402.15392} .
\bibitem[{Lin et~al.(2018)Lin, Stanley, Ghassemi and Nemati}]{40}
\bibinfo{author}{Lin, R.}, \bibinfo{author}{Stanley, M.D.}, \bibinfo{author}{Ghassemi, M.M.}, \bibinfo{author}{Nemati, S.}, \bibinfo{year}{2018}.
\newblock \bibinfo{title}{A deep deterministic policy gradient approach to medication dosing and surveillance in the icu}, in: \bibinfo{booktitle}{2018 40th Annual International Conference of the IEEE Engineering in Medicine and Biology Society (EMBC)}, \bibinfo{organization}{IEEE}. pp. \bibinfo{pages}{4927--4931}.
\bibitem[{Liu et~al.(2024a)Liu, Xie, Shu, Chen, Sun, Zhong, Liang, Li, Yang, Han et~al.}]{15}
\bibinfo{author}{Liu, J.}, \bibinfo{author}{Xie, Y.}, \bibinfo{author}{Shu, X.}, \bibinfo{author}{Chen, Y.}, \bibinfo{author}{Sun, Y.}, \bibinfo{author}{Zhong, K.}, \bibinfo{author}{Liang, H.}, \bibinfo{author}{Li, Y.}, \bibinfo{author}{Yang, C.}, \bibinfo{author}{Han, Y.}, et~al., \bibinfo{year}{2024}a.
\newblock \bibinfo{title}{Value function assessment to different rl algorithms for heparin treatment policy of patients with sepsis in icu}.
\newblock \bibinfo{journal}{Artificial Intelligence in Medicine} \bibinfo{volume}{147}, \bibinfo{pages}{102726}.
\bibitem[{Liu et~al.(2024b)Liu, Xie, Shu, Chen, Sun, Zhong, Liang, Li, Yang, Han et~al.}]{41}
\bibinfo{author}{Liu, J.}, \bibinfo{author}{Xie, Y.}, \bibinfo{author}{Shu, X.}, \bibinfo{author}{Chen, Y.}, \bibinfo{author}{Sun, Y.}, \bibinfo{author}{Zhong, K.}, \bibinfo{author}{Liang, H.}, \bibinfo{author}{Li, Y.}, \bibinfo{author}{Yang, C.}, \bibinfo{author}{Han, Y.}, et~al., \bibinfo{year}{2024}b.
\newblock \bibinfo{title}{Value function assessment to different rl algorithms for heparin treatment policy of patients with sepsis in icu}.
\newblock \bibinfo{journal}{Artificial Intelligence in Medicine} \bibinfo{volume}{147}, \bibinfo{pages}{102726}.
\bibitem[{Mihaylova et~al.(2012)Mihaylova, Emberson, Blackwell, Keech, Simes, Barnes, Voysey, Gray, Collins, Baigent et~al.}]{8}
\bibinfo{author}{Mihaylova, B.}, \bibinfo{author}{Emberson, J.}, \bibinfo{author}{Blackwell, L.}, \bibinfo{author}{Keech, A.}, \bibinfo{author}{Simes, J.}, \bibinfo{author}{Barnes, E.}, \bibinfo{author}{Voysey, M.}, \bibinfo{author}{Gray, A.}, \bibinfo{author}{Collins, R.}, \bibinfo{author}{Baigent, C.}, et~al., \bibinfo{year}{2012}.
\newblock \bibinfo{title}{The effects of lowering ldl cholesterol with statin therapy in people at low risk of vascular disease: meta-analysis of individual data from 27 randomised trials.}
\newblock \bibinfo{journal}{Lancet (London, England)} \bibinfo{volume}{380}, \bibinfo{pages}{581--590}.
\bibitem[{Nemati et~al.(2016)Nemati, Ghassemi and Clifford}]{14}
\bibinfo{author}{Nemati, S.}, \bibinfo{author}{Ghassemi, M.M.}, \bibinfo{author}{Clifford, G.D.}, \bibinfo{year}{2016}.
\newblock \bibinfo{title}{Optimal medication dosing from suboptimal clinical examples: A deep reinforcement learning approach}, in: \bibinfo{booktitle}{2016 38th annual international conference of the IEEE engineering in medicine and biology society (EMBC)}, \bibinfo{organization}{IEEE}. pp. \bibinfo{pages}{2978--2981}.
\bibitem[{Ng et~al.(2000)Ng, Russell et~al.}]{16}
\bibinfo{author}{Ng, A.Y.}, \bibinfo{author}{Russell, S.}, et~al., \bibinfo{year}{2000}.
\newblock \bibinfo{title}{Algorithms for inverse reinforcement learning.}, in: \bibinfo{booktitle}{Icml}, p.~\bibinfo{pages}{2}.
\bibitem[{Piot et~al.(2014)Piot, Geist and Pietquin}]{45}
\bibinfo{author}{Piot, B.}, \bibinfo{author}{Geist, M.}, \bibinfo{author}{Pietquin, O.}, \bibinfo{year}{2014}.
\newblock \bibinfo{title}{Boosted and reward-regularized classification for apprenticeship learning}, in: \bibinfo{booktitle}{Proceedings of the 2014 international conference on Autonomous agents and multi-agent systems}, pp. \bibinfo{pages}{1249--1256}.
\bibitem[{Qiu et~al.(2022)Qiu, Tan, Li, Chen, Ru and Jin}]{12}
\bibinfo{author}{Qiu, X.}, \bibinfo{author}{Tan, X.}, \bibinfo{author}{Li, Q.}, \bibinfo{author}{Chen, S.}, \bibinfo{author}{Ru, Y.}, \bibinfo{author}{Jin, Y.}, \bibinfo{year}{2022}.
\newblock \bibinfo{title}{A latent batch-constrained deep reinforcement learning approach for precision dosing clinical decision support}.
\newblock \bibinfo{journal}{Knowledge-based systems} \bibinfo{volume}{237}, \bibinfo{pages}{107689}.
\bibitem[{Ribba et~al.(2020)Ribba, Dudal, Lav{\'e} and Peck}]{9}
\bibinfo{author}{Ribba, B.}, \bibinfo{author}{Dudal, S.}, \bibinfo{author}{Lav{\'e}, T.}, \bibinfo{author}{Peck, R.W.}, \bibinfo{year}{2020}.
\newblock \bibinfo{title}{Model-informed artificial intelligence: reinforcement learning for precision dosing}.
\newblock \bibinfo{journal}{Clinical Pharmacology \& Therapeutics} \bibinfo{volume}{107}, \bibinfo{pages}{853--857}.
\bibitem[{Schamberg et~al.(2022)Schamberg, Badgeley, Meschede-Krasa, Kwon and Brown}]{11}
\bibinfo{author}{Schamberg, G.}, \bibinfo{author}{Badgeley, M.}, \bibinfo{author}{Meschede-Krasa, B.}, \bibinfo{author}{Kwon, O.}, \bibinfo{author}{Brown, E.N.}, \bibinfo{year}{2022}.
\newblock \bibinfo{title}{Continuous action deep reinforcement learning for propofol dosing during general anesthesia}.
\newblock \bibinfo{journal}{Artificial Intelligence in Medicine} \bibinfo{volume}{123}, \bibinfo{pages}{102227}.
\bibitem[{Siegel et~al.(2021)Siegel, Miller and Jemal}]{7}
\bibinfo{author}{Siegel, R.}, \bibinfo{author}{Miller, K.}, \bibinfo{author}{Jemal, R.A.}, \bibinfo{year}{2021}.
\newblock \bibinfo{title}{Cancer facts \& figures 2021.} .
\bibitem[{Smith et~al.(2023)Smith, Khojandi, Vasudevan, Shafi and Davis}]{13}
\bibinfo{author}{Smith, B.}, \bibinfo{author}{Khojandi, A.}, \bibinfo{author}{Vasudevan, R.}, \bibinfo{author}{Shafi, N.}, \bibinfo{author}{Davis, R.}, \bibinfo{year}{2023}.
\newblock \bibinfo{title}{Uncovering bias in reinforcement learning for heparin treatment planning} .
\bibitem[{Sutton(1990)}]{29}
\bibinfo{author}{Sutton, R.S.}, \bibinfo{year}{1990}.
\newblock \bibinfo{title}{Integrated architectures for learning, planning, and reacting based on approximating dynamic programming}, in: \bibinfo{booktitle}{Machine learning proceedings 1990}. \bibinfo{publisher}{Elsevier}, pp. \bibinfo{pages}{216--224}.
\bibitem[{Wadelius and Pirmohamed(2007)}]{1}
\bibinfo{author}{Wadelius, M.}, \bibinfo{author}{Pirmohamed, M.}, \bibinfo{year}{2007}.
\newblock \bibinfo{title}{Pharmacogenetics of warfarin: current status and future challenges}.
\newblock \bibinfo{journal}{The pharmacogenomics journal} \bibinfo{volume}{7}, \bibinfo{pages}{99--111}.
\bibitem[{Wulfmeier et~al.(2015)Wulfmeier, Ondruska and Posner}]{32}
\bibinfo{author}{Wulfmeier, M.}, \bibinfo{author}{Ondruska, P.}, \bibinfo{author}{Posner, I.}, \bibinfo{year}{2015}.
\newblock \bibinfo{title}{Maximum entropy deep inverse reinforcement learning}.
\newblock \bibinfo{journal}{arXiv preprint arXiv:1507.04888} .
\bibitem[{Yu et~al.(2019a)Yu, Liu and Zhao}]{46}
\bibinfo{author}{Yu, C.}, \bibinfo{author}{Liu, J.}, \bibinfo{author}{Zhao, H.}, \bibinfo{year}{2019}a.
\newblock \bibinfo{title}{Inverse reinforcement learning for intelligent mechanical ventilation and sedative dosing in intensive care units}.
\newblock \bibinfo{journal}{BMC medical informatics and decision making} \bibinfo{volume}{19}, \bibinfo{pages}{111--120}.
\bibitem[{Yu et~al.(2019b)Yu, Ren and Liu}]{44}
\bibinfo{author}{Yu, C.}, \bibinfo{author}{Ren, G.}, \bibinfo{author}{Liu, J.}, \bibinfo{year}{2019}b.
\newblock \bibinfo{title}{Deep inverse reinforcement learning for sepsis treatment}, in: \bibinfo{booktitle}{2019 IEEE international conference on healthcare informatics (ICHI)}, \bibinfo{organization}{IEEE}. pp. \bibinfo{pages}{1--3}.
\bibitem[{Yu et~al.(2021)Yu, Kumar, Rafailov, Rajeswaran, Levine and Finn}]{30}
\bibinfo{author}{Yu, T.}, \bibinfo{author}{Kumar, A.}, \bibinfo{author}{Rafailov, R.}, \bibinfo{author}{Rajeswaran, A.}, \bibinfo{author}{Levine, S.}, \bibinfo{author}{Finn, C.}, \bibinfo{year}{2021}.
\newblock \bibinfo{title}{Combo: Conservative offline model-based policy optimization}.
\newblock \bibinfo{journal}{Advances in neural information processing systems} \bibinfo{volume}{34}, \bibinfo{pages}{28954--28967}.
\bibitem[{Yu et~al.(2020)Yu, Thomas, Yu, Ermon, Zou, Levine, Finn and Ma}]{28}
\bibinfo{author}{Yu, T.}, \bibinfo{author}{Thomas, G.}, \bibinfo{author}{Yu, L.}, \bibinfo{author}{Ermon, S.}, \bibinfo{author}{Zou, J.Y.}, \bibinfo{author}{Levine, S.}, \bibinfo{author}{Finn, C.}, \bibinfo{author}{Ma, T.}, \bibinfo{year}{2020}.
\newblock \bibinfo{title}{Mopo: Model-based offline policy optimization}.
\newblock \bibinfo{journal}{Advances in Neural Information Processing Systems} \bibinfo{volume}{33}, \bibinfo{pages}{14129--14142}.
\bibitem[{Yue et~al.(2023)Yue, Wang, Shao, Zhang, Lin, Ren and Zhang}]{50}
\bibinfo{author}{Yue, S.}, \bibinfo{author}{Wang, G.}, \bibinfo{author}{Shao, W.}, \bibinfo{author}{Zhang, Z.}, \bibinfo{author}{Lin, S.}, \bibinfo{author}{Ren, J.}, \bibinfo{author}{Zhang, J.}, \bibinfo{year}{2023}.
\newblock \bibinfo{title}{Clare: Conservative model-based reward learning for offline inverse reinforcement learning}.
\newblock \bibinfo{journal}{arXiv preprint arXiv:2302.04782} .
\bibitem[{Zeng et~al.(2022)Zeng, Li, Garcia and Hong}]{34}
\bibinfo{author}{Zeng, S.}, \bibinfo{author}{Li, C.}, \bibinfo{author}{Garcia, A.}, \bibinfo{author}{Hong, M.}, \bibinfo{year}{2022}.
\newblock \bibinfo{title}{Maximum-likelihood inverse reinforcement learning with finite-time guarantees}.
\newblock \bibinfo{journal}{Advances in Neural Information Processing Systems} \bibinfo{volume}{35}, \bibinfo{pages}{10122--10135}.
\bibitem[{Zeng et~al.(2023)Zeng, Li, Garcia and Hong}]{51}
\bibinfo{author}{Zeng, S.}, \bibinfo{author}{Li, C.}, \bibinfo{author}{Garcia, A.}, \bibinfo{author}{Hong, M.}, \bibinfo{year}{2023}.
\newblock \bibinfo{title}{When demonstrations meet generative world models: A maximum likelihood framework for offline inverse reinforcement learning}.
\newblock \bibinfo{journal}{Advances in Neural Information Processing Systems} \bibinfo{volume}{36}, \bibinfo{pages}{65531--65565}.
\bibitem[{Ziebart et~al.(2008)Ziebart, Maas, Bagnell, Dey et~al.}]{31}
\bibinfo{author}{Ziebart, B.D.}, \bibinfo{author}{Maas, A.L.}, \bibinfo{author}{Bagnell, J.A.}, \bibinfo{author}{Dey, A.K.}, et~al., \bibinfo{year}{2008}.
\newblock \bibinfo{title}{Maximum entropy inverse reinforcement learning.}, in: \bibinfo{booktitle}{Aaai}, \bibinfo{organization}{Chicago, IL, USA}. pp. \bibinfo{pages}{1433--1438}.

\end{thebibliography}

\end{document}